\documentclass[sigconf]{acmart}

\usepackage{booktabs} 

\copyrightyear{2018} 
\acmYear{2018} 
\setcopyright{acmcopyright}
\acmConference[GECCO '18]{Genetic and Evolutionary Computation Conference}{July 15--19, 2018}{Kyoto, Japan}
\acmBooktitle{GECCO '18: Genetic and Evolutionary Computation Conference, July 15--19, 2018, Kyoto, Japan}
\acmPrice{15.00}
\acmDOI{10.1145/3205455.3205477}
\acmISBN{978-1-4503-5618-3/18/07}

\graphicspath{{images/}}
\usepackage{amsfonts,amsmath,amssymb}
\usepackage{graphicx} 
\usepackage{epstopdf} 
\usepackage{multirow,array}
\usepackage{enumitem}
\usepackage{color,soul}
\usepackage{mwe}

\usepackage[ruled]{algorithm2e}
\SetCommentSty{mycommfont}

\usepackage{bbm} 

\newcommand{\bbR}{\mathbb{R}}

\newcommand{\be}{\mathbf{e}}

\newcommand{\bx}{\mathbf{x}}

\newcommand{\bxi}{\boldsymbol{\xi}}

\newcommand{\cA}{\mathcal{A}}

\newcommand{\cE}{\mathcal{E}}

\newcommand{\cK}{\mathcal{K}}

\newcommand{\cO}{\mathcal{O}}
\newcommand{\cP}{\mathcal{P}}

\newcommand{\cS}{\mathcal{S}}

\newcommand{\floor}[1]{\left\lfloor #1 \right\rfloor}

\newcommand{\norm}[1]{\lVert #1\rVert} 

\makeatletter
\def\Ddots{\mathinner{\mkern1mu\raise\p@
\vbox{\kern7\p@\hbox{.}}\mkern2mu
\raise4\p@\hbox{.}\mkern2mu\raise7\p@\hbox{.}\mkern1mu}}
\makeatother

\newcommand{\clusteralg}{Hill-Valley Clustering}
\newcommand{\alg}{HillVallEA}
\newcommand{\algFull}{Hill-Valley Evolutionary Algorithm}

\begin{document}

\title[Real-Valued Evolutionary MMO driven by {\clusteralg}]{Real-Valued Evolutionary Multi-Modal Optimization driven by {\clusteralg}}
\author{S.C. Maree}
\affiliation{
\institution{Academic Medical Center}
\country{Amsterdam, The Netherlands}}
\email{s.c.maree@amc.uva.nl}

\author{T. Alderliesten}
\affiliation{\institution{Academic Medical Center}
\country{Amsterdam, The Netherlands}}
\email{t.alderliesten@amc.uva.nl}

\author{D. Thierens}
\affiliation{\institution{Utrecht University}
\country{Utrecht, The Netherlands}}
\email{d.thierens@uu.nl}

\author{P.A.N. Bosman}
\affiliation{\institution{Centrum Wiskunde \& Informatica}
\country{Amsterdam, The Netherlands}}
\email{peter.bosman@cwi.nl}

\renewcommand{\shortauthors}{Maree et. al.}

\begin{abstract}
Model-based evolutionary algorithms (EAs) adapt an underlying search model to features of the problem at hand, such as the linkage between problem variables. The performance of EAs often deteriorates as multiple modes in the fitness landscape are modelled with a unimodal search model. The number of modes is however often unknown a priori, especially in a black-box setting, which complicates adaptation of the search model. In this work, we focus on models that can adapt to the multi-modality of the fitness landscape. Specifically, we introduce {\clusteralg}, a remarkably simple approach to adaptively cluster the search space in niches, such that a single mode resides in each niche. In each of the located niches, a core search algorithm is initialized to optimize that niche. Combined with an EA and a restart scheme, the resulting Hill-Valley EA ({\alg}) is compared to current state-of-the-art niching methods on a standard benchmark suite for multi-modal optimization. Numerical results in terms of the detected number of global optima show that, in spite of its simplicity, {\alg} is competitive within the limited budget of the benchmark suite, and shows superior performance in the long run.
\end{abstract}

%
%



\maketitle

\section{Introduction}
Model-based evolutionary algorithms (EAs) adapt an underlying search model based on certain features of the fitness landscape. Classically, these features are related to linkage, or dependence, of problem variables. In this work, we focus on adaptive methods to exploit multi-modality of the fitness landscape. 

In this work, In optimization, a \textit{niche} is a subset of the search space where only one mode resides. When the fitness landscape is multi-modal, the performance of many EAs deteriorates as high-quality solutions can be found in different parts of the search space, which may prevent narrowing down the search to a specific region. Being able to explicitly deal with multi-modality may therefore be beneficial to many EAs. In addition, exploring multiple niches can provide additional insight into the structure of the problem at hand. Real-world problems are often not unimodal, and by providing the decision maker with multiple high-quality solutions, the final solution can be chosen based on external factors that are best considered once the set of interesting alternatives is known.

Niching methods originated as a tool for improving population diversity in EAs \cite{li17}, but are now generally designed for multi-modal optimization (MMO). MMO aims at locating multiple high-quality locally optimal solutions in a single run. One of the difficulties with niching methods is that they often introduce additional \textit{niching parameters} such as a minimal niche size or the number of niches \cite{li17}. The number of niches is however generally unknown a priori, and niching methods should therefore make few assumptions on their size, shape, or number. In this work, we therefore introduce a niching method based on a remarkably simple clustering method that is adaptive to the fitness landscape and investigate to what extent it can improve the multi-modal optimization power of existing well-known EAs. 

The remainder of the paper is organized as follows. In Section~\ref{sec:related work}, we discuss different niching concepts of related methods. In Section~\ref{sec:clustering}, we introduce the clustering method {\clusteralg}. In Section~\ref{sec:alg}, we describe how we combined clustering with a core search algorithm and a restart scheme into the {\algFull} ({\alg}). In Section~\ref{sec:experiments}, the performance of {\alg} is numerically evaluated on the CEC2013 niching benchmark suite \cite{CEC2013NichingCompetition}, that was also used for the GECCO'17 niching competition. Performance is compared to state-of-the-art niching methods. We finally summarize and discuss the results in Section~\ref{sec:Discussion} and conclude in Section~\ref{sec:conclusion}.

\section{Related Work}
\label{sec:related work}
Two-stage niching methods are a class of niching methods where the first stage is aimed at locating different niches in the search space. In the second stage, core search algorithms are initialized in the niches. The Nearest-better EA (NEA2+) \cite{preuss10, preuss12} is a real-valued two-phase MMO EA, which uses nearest better clustering (NBC) in the first phase to cluster the initial population. In NBC, a nearest better tree is constructed, i.e., a spanning tree that connects each solution to its nearest solution that has better fitness. By the definition of a local optimum, all nearby solutions are of worse fitness, and the outgoing edge of a local optimum is therefore expected to be longer than average. By removing long edges from the spanning tree, niches can be detected, and a core search algorithm then performs optimization in each niche. 
To determine edges that are longer than expected, two \textit{cutting rules} were tuned to a set of benchmark problems. When sampling few solutions with respect to the size of the search space, which occurs especially in high-dimensional problems, the distance between neighbouring solutions is heavily subject to randomness, and distinguishing long edges becomes hard. To reduce the dependency of the spread of the initial sample on the performance of the algorithm, the initial samples can be spread out more evenly over the search space compared to the commonly used uniform sampling  , by means of rejection sampling \cite{wessing16}. This is used in NEA2+ for a sharper tuning of the cutting rules, improving its performance.

In \cite{maree17}, an EA is introduced where the same nearest better tree is constructed as in NBC, but instead of cutting rules, the correlation between the search space and fitness values is exploited to determine the edges that need to be cut. This correlation is computed based on a set of fitted Gaussian mixture models, but the performance of this approach deteriorates in situations where Gaussian mixture models are ill-fitted.

Winner of the GECCO'17 niching competition is the repelling-subpopulations (RS-CMSA) algorithm \cite{ahrari16,ahrari17}. In RS-CMSA, instances of the core search algorithm CMSA are randomly initialized and maintain a minimum distance to each other, using rejection sampling. When all core search algorithms are terminated, the method is restarted, and the core search algorithms are furthermore kept away from previously located global optima. In this way, easy optima are found first, and the search is later pushed to unexplored regions of the search space. Rejection regions from located optima, referred to as \textit{taboo} regions, are considered to be (hyper-)spheres around the located optimum. The radius of the taboo region grows adaptively if the corresponding optimum is located multiple times, and shrinks again if rejection sampling fails too often. These adaptation parameters and the number of simultaneously run core search algorithms are then validated on the same set of benchmark problems used in this work.

One of the performance measures in the niching competition is \textit{precision}, meant as the fraction of relevant detected solutions compared to the overall number of detected solutions. In order to filter out the local optima, RS-CMSA applies a post-processing step based on the \textit{Hill-Valley} test \cite{ursem99}. The Hill-Valley test can be used to detect whether two solutions belong to the same niche, see Algorithm~\ref{alg:hill_valley}. To achieve this, an edge is drawn between the two solutions in the search space, and the fitness is evaluated on $N_t$ equidistantly located test points along the edge. If the fitness on any of the test points is worse than the fitness of both solutions, there is a \textit{hill} between the two solutions, and the two solutions presumably belong to different \textit{valleys} (niches).


\begin{algorithm}
\SetKwInOut{Function}{function}
\SetKwInOut{Input}{input}
\SetKwInOut{Output}{output}
\Function{$[B] = $Hill-Valley($\bx_\text{left}, \bx_\text{right},N_t$)}
\Input{Solutions $\bx_\text{left}, \bx_\text{right}$,\\ Number of test points $N_t$,\\ Fitness function $f$ \small{(to be minimized)}}
\Output{$\bx_\text{left}$ and $\bx_\text{right}$ belong to the same niche? \small{(boolean)}}
\BlankLine
\For{$k = 1,\ldots,N_t$ }
{
$\bx_\text{test} = \bx_\text{right} + \frac{k}{N_t + 1}(\bx_\text{left}-\bx_\text{right})$\;
\If{$\max\{f(\bx_\text{right}),f(\bx_\text{left})\} < f(\bx_\text{test})$}{
return false\;
}
}
return true\;
\caption{Hill-Valley test \cite{ursem99}}
\label{alg:hill_valley}
\end{algorithm}

The Hill-Valley test was applied in the multi-national EA \cite{ursem99} to each newly-sampled solution to determine if it belongs to an existing population, if two populations need to be merged, or if a new population needs to be formed. This approach was tested on noise-free two-dimensional problems while using five test points ($N_t = 5$). In RS-CMSA, the equidistant test points are replaced by a golden section search, with a maximum of $N_t = 10$ test points. More test points could be used due to its limited application only as post-processing step to filter out indistinct global optima \cite{ahrari16}.

\section{\clusteralg}
\label{sec:clustering}
In this work, we define multi-modal optimization (MMO) as the search for only global optima. In the presence of noise, or in real-world applications, one might however also be interested in locating high-quality local optima. Guided by the problems in the CEC2013 niching benchmark suite, the niching method we present in this work discards these local optima,  but this can be easily adapted.

Given a real-valued search space $X\subseteq\bbR^d$ and objective function $f : X \rightarrow \bbR$, that, without loss of generality, needs to be minimized, the aim of MMO is to find all $\bx^\star \in X$ that obtain the minimum possible objective value, i.e., 
\begin{equation}
f(\bx^\star) \leq f(\bx), \quad \forall \bx \in X \subseteq \bbR^d.
\end{equation}

The core of {\clusteralg} is the Hill-Valley test, combined with the concept of the nearest better tree. The idea is that solutions belonging to the same niche should be in the same cluster.

{\clusteralg} is initialized with a population $\cP$ of $|\cP| = N$ solutions, and these solutions are sorted such that the best solution is first. This best solution $\bx_0$ forms the first cluster $C_0 = \{\bx_0\}$. Then, we consider the second-best solution $\bx_1$, and the Hill-Valley test in Algorithm~\ref{alg:hill_valley} is used to test whether it belongs to the same niche as $\bx_0$. When it does, it is added to the cluster of $\bx_0$, denoted by $C_{(\bx_0)} = C_{(\bx_0)}\cup \{\bx_1\}$. Otherwise, as there are no other solutions with higher fitness to check against, a new cluster is formed, i.e., $C_1 = \{\bx_1\}$.

Next, the third-best solution $\bx_2$ is tested against the nearest solution that has better fitness, which can either be $\bx_0$ or $\bx_1$, depending on which one is nearer. If $\bx_2$ does not belong to the same niche as its nearest better solution, we test it against the second-nearest better solution. If it also does not belong to that niche, we create a new cluster from $\bx_2$. Multiple neighbours are checked as the nearest better solution could belong to a different niche, but a second-nearest better solution, which is located slightly further, but in a different direction, might belong to the same niche. For each solution, we check its $\gamma = d + 1$ nearest better neighbours, where $d$ is the problem dimensionality. In this way, even for the one-dimensional problems, at least two neighbours are considered. 

If a solution does not belong to any of the niches of its nearest neighbour, it forms a new cluster. This procedure is repeated for all solutions in the selection. 

The output is a set of clusters, each consisting of at least one solution. With these clusters, core search algorithms are initialized. 

To slightly reduce the number of function evaluations, we only check nearest better solutions from each cluster once. If two nearest better solutions belong to the same cluster, we only check the first, and if it is rejected, we reject the second without checking. 

\begin{algorithm}
\SetKwInOut{Function}{function}
\SetKwInOut{Input}{input}
\SetKwInOut{Output}{output}
\SetKwData{Sort}{Sort}
\SetKwData{Find}{Find}
\SetKwData{Add}{Add}
\Function{[$\cK$] = HillValleyClustering($\cS$)}
\Input{Set of solutions $\cS$,\\ search space volume $V_X$,\\ problem dimension $d$}
\Output{Set of clusters $\cK = \{C_0, C_1,\ldots,C_{K-1}\}$}
\BlankLine

\Sort solutions $\bx_i\in\cS$ on fitness value, fittest first\;

\BlankLine

\For{$i = 1,\ldots,|\cS|-1$}
{

\BlankLine
	\For{$j = 0,\ldots,i-1$}
	{
	$\delta_j \leftarrow \norm{\bx_i - \bx_j}$\tcp*{Euclidean distances to better $\bx_j$}
	}

	\BlankLine
	\For{$j = 0,\ldots,\min\{i-1,d\}$}
	{
	\BlankLine
		$k$ = index of the $j$-th nearest better solution in $\{\delta_j\}$;
		
		$N_t = 1 + \floor{\delta_{k} / \sqrt[d]{{V_X}/{|\cS|}}}$\;
		\BlankLine
		\If{not already checked cluster $C_{(\bx_k)}$}
		{		
		
		\If{Hill-Valley($\bx_{k}, \bx_i, N_t$)}
		{
			$C_{(\bx_k)} = C_{(\bx_k)} \cup \{\bx_i\}$\tcp*{add to cluster of $\bx_{k}$}
			break\;
		}
		}
	}
	
	\If{$\bx_i$ was not added to any cluster}
	{
	$C := \{\bx_i\}$\tcp*{new cluster containing $\bx_i$}
	$\cK = \cK \cup C$\tcp*{add new cluster to the cluster set}
	}
}

return $\cK$\tcp*{return cluster set}
\caption{\clusteralg}
\label{alg:clustering}
\end{algorithm}

\subsection{Number of test points $N_t$}
We base the number of test points $N_t$ on the (Euclidean) length of the edge connecting the two solutions, and distribute the points equidistantly over the edge. We compute the expected edge length (EEL) as the distance between two solutions when they would have been scattered equidistantly in the search space, by,
\begin{equation}
\label{eqn:EEL}
EEL =  \sqrt[d]{{V_X}/{|\cS|} },
\end{equation}
where $V_X$ is the volume of the search space $X$. The longer an edge is, compared to the expected edge length, the more test points are sampled, according to,
\begin{equation}
N_t = N_t(\bx_\text{left},\bx_\text{right}) = 1 + \floor{\frac{\norm{\bx_\text{left} - x_\text{right}}}{\text{EEL}}},
\label{eqn:number_of_testpoints}
\end{equation}
where the fraction is the current Euclidean edge length divided by the expected edge length. Note that $N_t$ is the maximum number of test points that is evaluated for a single call of the Hill-Valley test. When two solutions belong to the same niche, all points will be evaluated. If not, the Hill-Valley test rejects the edge before all $N_t$ points are evaluated. 

Alternatively, if the volume of the search space cannot be easily determined, the expected edge length can be replaced by the average edge length. For each solution, find its nearest better solution and the corresponding edge length. Then, compute the average edge length over all edge lengths. If the nearest better distances are stored, the computational overhead is a negligible $\cO(|\cS|)$, compared to a complexity of $\cO(d|\cS|^2)$ of Algorithm~\ref{alg:clustering}.

\section{{\algFull}}
\label{sec:alg}
After clustering the initial population using {\clusteralg} described in Algorithm~\ref{alg:clustering}, for each cluster, a core search algorithm is initialized. The niching methods discussed in this work all used different core search algorithms, which disguises the added value of a specific niching approach. We therefore equip {\alg} with different core search algorithms. 

\subsection{Core search algorithms}
We test {\alg} with the core search algorithms $\cA$ listed in Table~\ref{tab:popsizes}, which are CMSA \cite{beyer08} and different versions of AMaLGaM \cite{bosman13}. These algorithms are based on a Gaussian distribution, initialized by a mean $\mu$ and covariance matrix $\Sigma$, population size $N_c$, and return a single solution, i.e.,
\begin{equation}
[\bx] = \cA(\mu,\Sigma, N_c)
\end{equation} AMaLGaM is an estimation of distribution algorithm, where a Gaussian is fitted to selected solutions with maximum likelihood. 
CMSA was used instead of the more common CMA-ES \cite{hansen05}, as it was suggested to perform better when adding elitism \cite{ahrari16}. CMSA fits a Gaussian using the population mean of the previous generation. 

The methods have different recommended population sizes, see Table~\ref{tab:popsizes}. 
Likely, a smaller core population size is better, since the complexity of the problem within each niche is expected to be smaller than the complexity of the full problem.
Therefore, a small population might be sufficient to converge to the optimum within a niche, using fewer function evaluations. This would allow for generations of the core search algorithms to be run, which is beneficial especially when many clusters are detected.

\begin{table}
\caption{Different core search algorithms $\cA$ with corresponding recommended cluster (population) size $N_c^{\text{rec}}$ as taken from literature, for problem dimensionality $d$.}
\label{tab:popsizes}
\begin{tabular}{lcc}
\hline
Core search algorithm $\cA$ & abbr. & $N_c^{\text{rec}}$ \\
\hline
CMSA \cite{beyer08} & CMSA & $3\log d$ \\
AMaLGaM \cite{bosman13} & AM  & $17 + 3\cdot d\sqrt{d}$ \\
AMaLGaM-Univariate \cite{bosman13}& AMu & $10 \sqrt{d}$ \\
iAMaLGaM \cite{bosman13} &iAM & $10\sqrt{d}$ \\
iAMaLGaM-Univariate\cite{bosman13}  & iAMu & $4\sqrt{d}$ \\
\hline
\end{tabular}
\end{table}

Core search algorithms are initialized from a cluster by setting the $\mu$ as the cluster mean, and $\Sigma$ as the sample covariance matrix of the solutions in the cluster with respect to $\mu$. If the cluster consists of fewer than $d+1$ solutions, only the diagonal of $\Sigma$ is estimated and all other entries are set to be zero. If a cluster consists of only one solution, the covariance matrix is initialized by the identity matrix, multiplied by $0.01\cdot\text{EEL}$, so that newly sampled solutions will be closer-by than its nearest better solutions in the initial population.

CMSA is terminated using the recommended criteria \cite{hansen05}, and parameters set as in RS-CMSA \cite{ahrari16}, that is, if the improvement in fitness value over the last $10 + \floor{30 d / N_c}$ generations is less than $TOL = 10^{-5}$. Tolerance $TOL$ corresponds to the desired accuracy of optima in the CEC2013 niching benchmark suite, and should be adapted if a different accuracy is required for the problem at hand. Furthermore, fail-safe termination criteria are added to terminate the core search algorithms if the standard deviation of solutions in the search space reaches machine accuracy ($10^{-15}$) and if the condition number of the covariance matrix is larger than $10^{14}$.

Termination is enforced for AMaLGaM-variant core search algorithms if the maximum standard deviation of solutions in the search space is too low ($10^{-12}$), or if the standard deviation of fitness values is too low ($10^{-12}$). 

Finally, the whole optimization is terminated if the budget, in terms of a maximum number of function evaluations specified in the benchmark, is reached.

\begin{algorithm}
\SetKwInOut{Function}{function}
\SetKwInOut{Input}{input}
\SetKwInOut{Output}{output}
\SetKwData{Sort}{Sort}
\SetKwData{Find}{Find}
\SetKwData{Add}{Add}
\Function{[$\cE$] = {\alg}($\cA$)}
\Input{Core search algorithm $\cA$, \\ problem dimension $d$}
\Output{Set of presumed global optima $\cE$}
\BlankLine

$N = 16 d$\tcp*{initial population size}
$N_c = N_c^{\text{rec}}{(\cA)}$ \tcp*{cluster size}
$\tau = \tau(\cA)$ \tcp*{$\tau = 0.5$ for $\cA= $ CMSA or $\tau = 0.35$ else}
$\cE = \{\}$\tcp*{elitist archive}
\BlankLine
\While{budget remaining}
{
\BlankLine
\tcp{First phase - locating niches}
$\cP = \text{uniformly\_sample}(N)$\tcp*{also evaluates solutions}

$\cP = \cP \cup \cE$\tcp*{add elites to population}

$\cS = \text{truncation\_selection}(\cP,\tau)$\;

$\cK = \text{HillValleyClustering}(\cS)$\;

\BlankLine

\tcp{Second phase - niche optimization}
\ForEach{$C_i \in \cK$}
{
	\BlankLine
	\If{best solution in $C$ is an elite} {\tcp{skip already-optimized niches}continue}

	\tcp{Run core search algorithm}
	$\bx_i = \cA(\mu(C_i),\Sigma(C_i), N_c)$
	\BlankLine
	\If{$\bx_i$ is a presumed distinct global optimum}
	{
	
	\If{for any $\be\in\cE$, $\bx_i + TOL < \be$}
	{
		\tcp{Empty the archive}
		$\cE = \{\}$
	}
	\tcp{Add to elitist archive}
	$\cE = \cE \cup \bx_i$
	}
}
\BlankLine
\If{no new solution was added to $\cE$}
{
$N\leftarrow 2N$\tcp*{increase initial population size}
$N_c \leftarrow 1.2 N_c$\tcp*{increase cluster size}
}

}
\caption{\algFull}
\label{alg:restart}
\end{algorithm}

\subsection{Restarts with an elitist archive}
There are two population size parameters that need to be set. First, there is the initial population size $N$ used as input for {\clusteralg}. The second parameter is the population size of the core search algorithms $N_c$. Schemes that grow the population of an EA over time, i.e., with restarts, are common to overcome setting these parameters, such as used in I-POP-CMA \cite{auger05}, the parameter-free genetic algorithm \cite{harik99}, and the interleaved multistart scheme \cite{bouter17}.

{\alg} is initialized with a population $\cP$ of size $N = 16d$, where $d$ is the problem dimensionality. To enhance performance, truncation selection is performed with a selection fraction $\tau$ based on the core search algorithm used. This is $\tau = 0.35$ for the versions of AMaLGaM and $\tau = 0.5$ for CMSA.  By performing selection, less effort is spent on low-fitness regions in the search space. On the other hand, small niches could be accidentally discarded when selection pressure is too high. 

The selection $\cS$ of $\floor{\tau N}$ solutions is then clustered using {\clusteralg}, of which pseudo code is given in Algorithm~\ref{alg:clustering}. From each of the resulting clusters, a core search algorithm is ran one after another. By construction of {\clusteralg}, the resulting clusters are sorted on the fitness values of the best, and the cluster containing the overall best solution is ran first. The run order is of importance when more clusters are found than the computational budget allows to run until convergence.
 
After running all core search algorithms with a cluster size of $N_c = N_c^\text{rec}$, the result is a set of $|\cK|$ candidate optima. A post-processing step is performed to discard the local optima and the duplicate global optima. First, all solutions that are more than $TOL$ worse than the all-time best, are discarded. Then, all remaining optima are tested for being in a different niche by the Hill-Valley test. The expected edge length does not make sense now, so we use the originally proposed $N_t = 5$ test points, similar to the post-processing step in \cite{ahrari16}. All presumed distinct global optima are then added to the elitist archive $\cE$.

If, after all core search algorithms are terminated, there is budget remaining, we restart the procedure. To prevent overhead, we add all elite solutions (the presumed distinct global optima) to the new population. If, after clustering, the cluster-best is any of these elites, we do not re-run that cluster, as that niche was presumably already optimized.

If no new global optima are found in a run, restarts are performed with increased population size to be able to detect smaller niches. Furthermore, the population size of the core search algorithms is also increased in order to find optima in more complex niches. 

The initial population size is increased with a factor $N^\text{inc} = 2$ after each run in which no new global optima are detected. The core search algorithm starts with the recommended value in Table~\ref{tab:popsizes}, and is incremented by a factor $N_c^\text{inc} = 1.2$. Pseudo code of {\alg} is given in Algorithm~\ref{alg:restart}.

\section{Experiments}
\label{sec:experiments}
We evaluate the performance of {\alg} on the test problems in the CEC2013 niching benchmark suite \cite{CEC2013NichingCompetition}. The benchmark consists of 20 problems, described in Table~\ref{tab:2dbenchmarks}, to be solved within the corresponding benchmark budget in terms of a limited number of function evaluations. 

For each of the benchmark problems, the location of the optima and the corresponding fitness values are known. As performance measure, the peak ratio is used, which measures the fraction of global optima detected, computed according to the guidelines of the niching benchmark suite \cite{CEC2013NichingCompetition}. In contrast to the full benchmark suite, where a range of accuracies $\varepsilon$ is used, and the final peak ratio is the mean of these results, we only use the highest accuracy of $ \varepsilon = 10^{-5}$. Due to the post-processing step in {\alg}, all local optima and duplicate global optima are filtered out, making the results independent of the choice of $\varepsilon \geq 10^{-5}$.

All benchmark functions are defined on a bounded domain, making computation of the search space volume in Equation~\eqref{eqn:EEL} straightforward.

All experiments in this work are repeated 50 times, and resulting performance measures are averaged over all repetitions. Results are tested for statistical significance with a two-sided 2-sample Wilcoxon rank sum test ($\alpha = 0.05$). A non-parametric test is used because the peak ratio is discrete, making it non-normally distributed.

For comparison, RS-CMSA was re-implemented in C++ based on the on-line available MATLAB implementation. Performance, measured in the peak ratio, was comparable to the results published in \cite{ahrari17}.

In Section~\ref{sec:restart_efficiency} we compare performance of different versions of the restart scheme. In Sections \ref{sec:restart_tuning} and \ref{sec:advanced_initial_sampling}, an attempt was made to further improve {\alg} by respectively tuning hyper-parameters and by incorporating an advanced sampling scheme. Then, in Section~\ref{sec:core_search_algorithms}, the performance of {\alg} with multiple core search algorithms is compared. Finally, in Section~\ref{sec:long_run}, performance is investigated on the benchmark suite with extended computational budget.

\begin{table}
\caption{Niching benchmark suite from the CEC2013 special session on multi-modal optimization \cite{CEC2013NichingCompetition}. For each problem the function name, problem dimensionality $d$, number of global optima $\#gopt$, and local optima $\#lopt$ and budget in terms of function evaluations are given.}
\label{tab:2dbenchmarks}
\small
\begin{tabular}{lccccc}
\toprule
\# & Function name & $d$ & \#gopt & \#lopt & budget \\
\toprule
1 & Five-Uneven-Peak Trap & 1 & 2 & 3 & 50K \\
2 & Equal Maxima & 1 & 5 & 0 & 50K \\
3 & Uneven Decreasing Maxima & 1 & 1 & 4 & 50K \\
4 & Himmelblau & 2 & 4 & 0 & 50K \\
5 & Six-Hump Camel Back & 2 & 2 & 5 & 50K \\
6 & Shubert & 2 & 18 & many & 200K \\
7 & Vincent & 2 & 36 & 0 & 200K\\
8 & Shubert & 3 & 81 & many & 400K\\
9 & Vincent & 3 & 216 & 0 & 400K\\
10 & Modified Rastrigin & 2 & 12 & 0 & 200K\\
11 & Composition Function 1 & 2 & 6 & many &200K\\
12 & Composition Function 2 & 2 & 8 &many & 200K\\
13 & Composition Function 3 & 2 & 6 &many & 200K\\
14 & Composition Function 3 & 3 & 6 &many & 400K\\
15 & Composition Function 4 & 3 & 8 &many & 400K\\
16 & Composition Function 3 & 5 & 6 &many & 400K\\
17 & Composition Function 4 & 5 & 8 &many & 400K\\
18 & Composition Function 3 & 10 & 6 & many &400K\\
19 & Composition Function 4 & 10 & 8 &many & 400K\\
20 & Composition Function 4 & 20 & 8 &many & 400K\\
\bottomrule
\end{tabular}
\end{table}

\subsection{Restart efficiency}
\label{sec:restart_efficiency}
In an attempt to reduce the computational overhead of the restart scheme, the located global optima are added to the population, as a representative for that entire niche. After clustering, these global niches are then skipped in the resulting local phase. Alternatively, both global and local optima could be added, which would prevent local niches from being re-explored as well. We compare these to the use of a restart scheme without addition of optima to the population for {\alg} and the simplest core search algorithm, AMaLGaM-Univariate and run it on all of the benchmark problems limited to the benchmark budget. Achieved peak ratios are given in Table~\ref{tab:restart_effectiveness}. 

Adding all optima to the population deteriorates performance significantly, especially notable for problems 6, 8, 13--20, all functions with many local optima. By adding all local optima back to the population, the population grows faster, and more of the budget is required for clustering these local optima over and over after each restart. Problems 9 has no local optima, so there is no difference between adding all located optima or only the located global optima. Compared to not adding optima, a huge improvement in peak ratio from 0.503 to 0.875 can be observed.

\begin{table}
\caption{Peak ratios of {\alg}-AMu with three forms of the restart scheme. Average (avg.) peak ratio is computed over all benchmark problems. The best achieved results per problem, and those results not statistically different from it, are typeset in boldface. The right three columns show the fraction of budget (in function evaluations) used by the `only global' restart scheme of HillVallEA-AMU for initialization (init), {\clusteralg} (HVC) and local optimization (lopt). }
\label{tab:restart_effectiveness}
\small
\begin{tabular}{cccc|ccc}
\toprule	
 & \multicolumn{3}{c|}{Peak ratio of Restart Schemes} & \multicolumn{3}{c}{Budget usage} \\						
\#	&	all optima	&	only global	&	no optima & init & HVC & lopt	\\
\toprule							
1 &\textbf{1.000} &\textbf{1.000} &\textbf{1.000}  &	0.04	&	0.05	&	0.91	\\
2 &\textbf{1.000} &\textbf{1.000} &\textbf{1.000}&	0.07	&	0.11	&	0.82	\\
3 &\textbf{1.000} &\textbf{1.000} &\textbf{1.000} &	0.04	&	0.06	&	0.90	\\
4 &\textbf{1.000} &\textbf{1.000} &\textbf{1.000} &	0.08	&	0.12	&	0.80	\\
5 &\textbf{1.000} &\textbf{1.000} &\textbf{1.000}&	0.03	&	0.04	&	0.93	\\
6 &0.942 &\textbf{0.997} &\textbf{0.999}&	0.01	&	0.02	&	0.97	\\
7 &\textbf{1.000} &\textbf{1.000} &0.969 &	0.23	&	0.36	&	0.41	\\
8 &0.263 &0.656 &\textbf{0.776}&	0.01	&	0.02	&	0.97	\\
9 &\textbf{0.875} &\textbf{0.875} &0.503 &	0.10	&	0.26	&	0.64	\\
10 &\textbf{1.000} &\textbf{1.000} &\textbf{1.000}&	0.03	&	0.07	&	0.90	\\
11 &\textbf{1.000} &\textbf{1.000} &\textbf{1.000}&	0.04	&	0.07	&	0.89\\
12 &\textbf{1.000} &\textbf{1.000} &\textbf{1.000}&	0.01	&	0.02	&	0.97\\
13 &0.857 &\textbf{0.973} &\textbf{0.953}&	0.02	&	0.03	&	0.95\\
14 &0.713 &\textbf{0.783} &\textbf{0.760}&	0.01	&	0.01	&	0.98\\
15 &0.723 &\textbf{0.750} &\textbf{0.750}&	0.01	&	0.01	&	0.98\\
16 &\textbf{0.670} &\textbf{0.673} &\textbf{0.670}&	0.01	&	0.01	&	0.98\\
17 &0.703 &\textbf{0.745} &\textbf{0.743}&	0.01	&	0.01	&	0.98\\
18 &\textbf{0.653} &\textbf{0.657} &\textbf{0.643}&	0.00	&	0.01	&	0.99\\
19 &\textbf{0.512} &\textbf{0.512} &0.490&	0.01	&	0.01	&	0.98\\
20 &\textbf{0.313} &\textbf{0.318} &\textbf{0.315}&	0.01	&	0.01	&	0.98\\
\bottomrule
avg. & 0.811 &  0.847 &  0.829 & 0.04 & 0.07 & 0.90\\
\bottomrule							
\end{tabular}
\end{table}

 \subsection{Parameter tuning of the restart scheme}
\label{sec:restart_tuning}
The restart scheme has four parameters: the initial population size $N$, the increment factor $N^\text{inc}$, the initial cluster size $N_c$ and the increment factor $N^\text{inc}_c$. These parameters have been tuned manually to intuitively make sense, and are found to perform well. To demonstrate robustness of the scheme, we tuned these parameters further in an automatic fashion. We set $N = 2^{\xi_0}\cdot d, N^\text{inc} = \xi_1, N_C = \xi_2\cdot N_C^\text{rec}$, $N_c^\text{inc} = \xi_3$, with $\xi_0 \in [4,10]$ and $\xi_i \in [1,5]$ for $i = 1,2,3$. The manually chosen parameters as stated in Algorithm~\ref{alg:restart} can then be written as $\bxi = (4,2,1,1.2)$.

\begin{figure}
\begin{center}
{\includegraphics[width=\columnwidth]{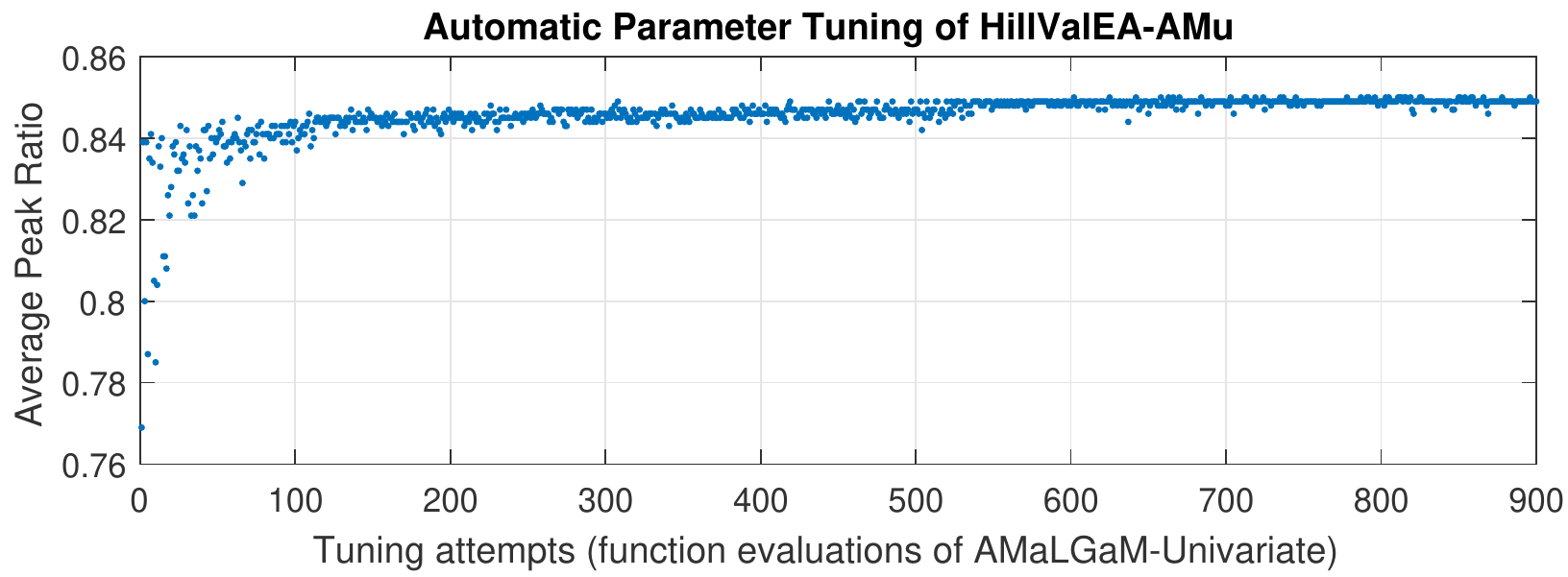}}
\caption{Automatic parameter tuning of the restart scheme of {\alg}-AMu on the average peak ratio of 50 repetitions of the full benchmark suite.}
\label{fig:tuning_attempts}
\end{center}
\end{figure}

AMaLGaM-Univariate was used to maximize the performance of {\alg}-AMu, measured by the average of achieved peak ratio over 50 repetitions of the 20 benchmark problems. The obtained average peak ratios are shown in Figure~\ref{fig:tuning_attempts}. The best obtained average peak ratio is 0.850, attained for the values $\bxi = (7.872, 3.839, 1.365, 1.351)$. This is only a slight improvement over the manually tuned parameters (0.847), but with very different values for the parameters.

\subsection{Advanced initial population sampling}
\label{sec:advanced_initial_sampling}
Both NEA2+ and RLSIS sample the initial population using the MaxiMin Reconstruction (MMR) method, where, based on rejection sampling, the initial population is more evenly distributed over the search space compared to uniform sampling or Latin hypercube sampling \cite{wessing16}.
We replaced $\cP=\text{uniformly\_sample}(N)$ in {\alg} (Algorithm~\ref{alg:restart}) by a sample generated with MMR. {\alg}-AMu+MMR achieved an average peak ratio of 0.849 over the entire benchmark set. This is not statistically significantly different from the previously obtained 0.847 with uniform sampling (see Table~\ref{alg:restart}). For individual problems, the differences in peak ratio were all less than 0.02, which is again not statistically significant. Standard uniform sampling is therefore preferred, as it is computationally cheaper.

\subsection{Core search algorithms in {\alg}}
\label{sec:core_search_algorithms}
We equip {\alg} with each of the core search algorithms in Table~\ref{tab:popsizes}. The top three performing competitors in the GECCO'17 niching competition, based on the CEC2013 benchmark suite, are RS-CMSA, with an average peak ratio of 0.856, RLSIS \cite{wessing16} with 0.822, and NEA2+ with 0.810 over the entire benchmark set.

Obtained peak ratios for each of the problems are given in Table~\ref{tab:results_localopts}. For all core search algorithms, {\alg} performs well, with the average peak ratio over all problems between 0.816--0.847, outperforming NEA2+ in all cases, and RLSIS for four different core search algorithms.

Problems 1--5 and 10 are easily solved within a fraction of the benchmark budget. {\alg} also solves problems 7, 11, and 12 fully, but this requires almost the entire budget. 

The largest differences can be observed for problems 8, 9, and 20. Problem 8 (Shubert function, 3D) has 81 global optima, all with a small niche. A strategy that allows many restarts will perform well here, which is the case with {\alg}-AMu, which maintains a cluster size and requires fewer function evaluations to converge each local optimizer. Problem 9 (Vincent function, 3D) has largely varying niche sizes, but no local optima. {\clusteralg} works very well here in detecting the large niches first, and the restart scheme prevents re-optimizing these niches, which significantly outperforms the peak ratio achieved by RS-CMSA. Problem 20 (CF4), has the largest dimensionality in the benchmark, $d = 20$, and has one very large niche. RS-CMSA achieves a better success rate than any {\alg} variant for this problem. 

\begin{table}
\caption{Peak ratio for RS-CMSA and {\alg} equipped with different core search algorithms, for all benchmark problems and with budget limited to the benchmark budget. Average (avg.) peak ratio is computed over all problems. The best achieved results per problem, and those results not statistically different from it, are typeset in boldface.}
\label{tab:results_localopts}
\small
\begin{tabular}{cc|ccccc}
\toprule													
	&	RS	& 	\multicolumn{5}{c}{\alg}									\\
\#	& 	CMSA	&	AM	&	AMu 	&	CMSA	&	iAM	&	iAMu	\\
\toprule													
1 &\textbf{1.000} &\textbf{1.000} &\textbf{1.000} &\textbf{1.000} &\textbf{1.000} &\textbf{1.000}\\
2 &\textbf{1.000} &\textbf{1.000} &\textbf{1.000} &\textbf{1.000} &\textbf{1.000} &\textbf{1.000}\\
3 &\textbf{1.000} &\textbf{1.000} &\textbf{1.000} &\textbf{1.000} &\textbf{1.000} &\textbf{1.000}\\
4 &\textbf{1.000} &\textbf{1.000} &\textbf{1.000} &\textbf{0.999} &\textbf{1.000} &\textbf{1.000}\\
5 &\textbf{1.000} &\textbf{1.000} &\textbf{1.000} &\textbf{1.000} &\textbf{1.000} &\textbf{1.000}\\
6 &\textbf{1.000} &0.976 &\textbf{0.997} &0.980 &0.989 &0.976\\
7 &0.996 &\textbf{1.000} &\textbf{1.000} &\textbf{1.000} &\textbf{1.000} &\textbf{1.000}\\
8 &\textbf{0.881} &0.410 &0.656 &0.437 &0.465 &0.501\\
9 &0.721 &0.782 &0.875 &0.882 &0.851 &\textbf{0.922}\\
10 &\textbf{1.000} &\textbf{1.000} &\textbf{1.000} &\textbf{1.000} &\textbf{1.000} &\textbf{1.000}\\
11 &\textbf{1.000} &\textbf{1.000} &\textbf{1.000} &\textbf{1.000} &\textbf{1.000} &\textbf{1.000}\\
12 &\textbf{1.000} &\textbf{1.000} &\textbf{1.000} &\textbf{1.000} &\textbf{1.000} &\textbf{1.000}\\
13 &\textbf{1.000} &0.923 &0.973 &0.973 &0.957 &0.963\\
14 &\textbf{0.780} &\textbf{0.763} &\textbf{0.783} &\textbf{0.793} &0.750 &\textbf{0.783}\\
15 &\textbf{0.748} &0.740 &\textbf{0.750} &0.728 &\textbf{0.740} &0.735\\
16 &\textbf{0.667} &\textbf{0.670} &\textbf{0.673} &\textbf{0.670} &\textbf{0.677} &\textbf{0.677}\\
17 &0.688 &0.720 &\textbf{0.745} &0.670 &\textbf{0.720} &0.720\\
18 &\textbf{0.667} &0.623 &\textbf{0.657} &0.637 &0.640 &\textbf{0.667}\\
19 &\textbf{0.505} &0.465 &\textbf{0.512} &\textbf{0.512} &0.490 &\textbf{0.505}\\
20 &\textbf{0.465} &0.238 &0.318 &0.338 &0.275 &0.315\\
\bottomrule
avg. & 0.856 &  0.816 &  0.847 &  0.831 &  0.829 &  0.838 \\
\bottomrule													
\end{tabular}
\end{table}

\subsection{Maximum Achievable Peak Ratio}
\begin{figure*}
\begin{center}
{\includegraphics[trim=2.6cm 0.5cm 3.2cm 1.4cm, width=\textwidth, clip]{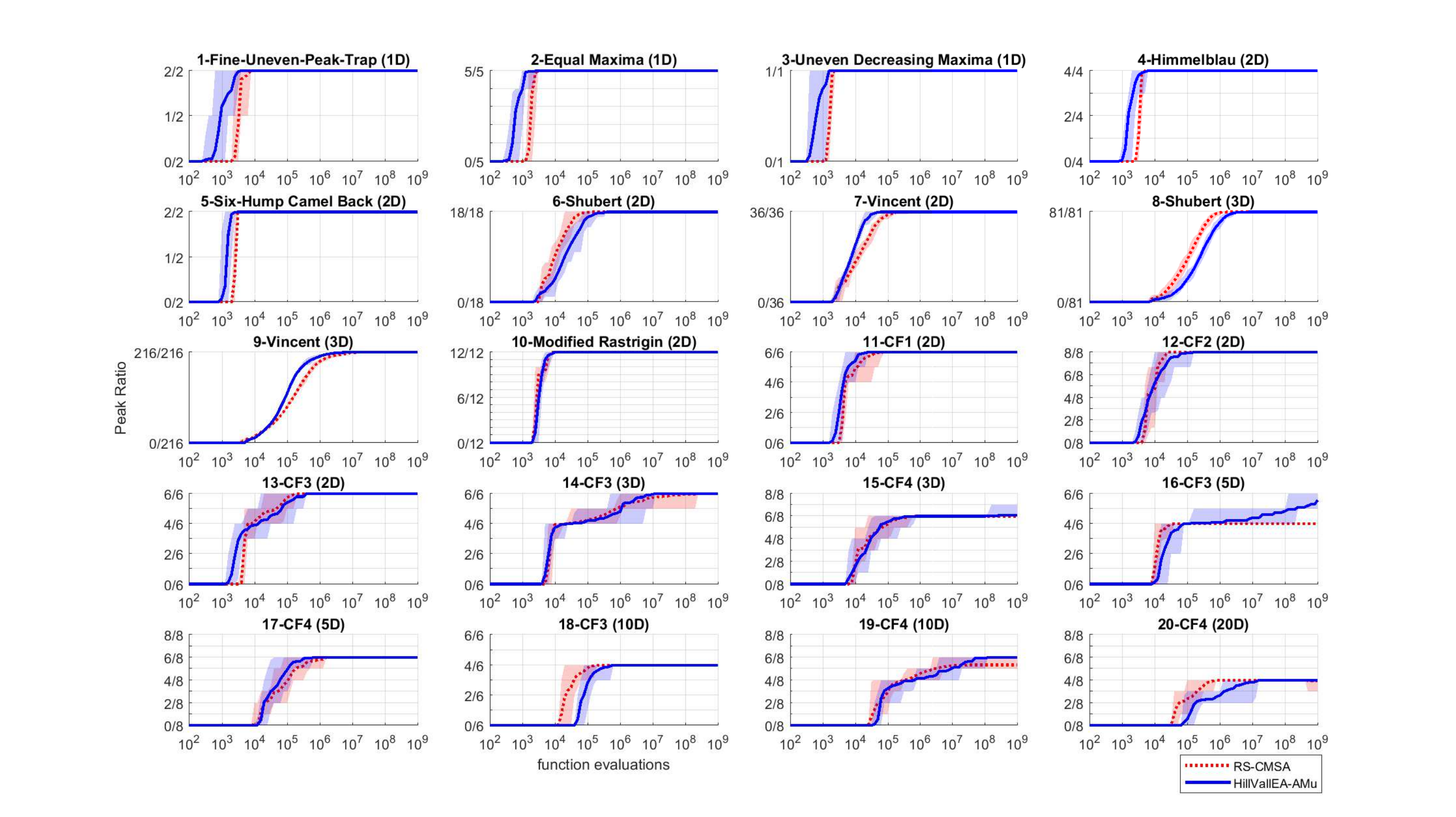}}
\caption{For each benchmark problem, the peak ratio is shown as a function of the number of function evaluations, for RS-CMSA and {\alg}-AMu, with an increased budget of 50 times the original benchmark budget. Shaded areas show the min--max range over 20 repetitions of the experiment.}
\label{fig:long_budget}
\end{center}
\end{figure*}

The benchmark budget is tight, as not all problems can be solved within the budget. We therefore increase the computational budget to one billion ($10^9$) function evaluations. Due to computational limits, experiments are repeated 20 times (instead of 50 in the other experiments). We consider the peak ratio for each individual problem in the benchmark versus the number of function evaluations. In Figure~\ref{fig:long_budget} the results for {\alg}-AMu and RS-CMSA are given.

Problems 1--5 and 10 are solved easily within the original benchmark budget. On the Vincent function, (7 and 9), {\alg}-AMu outperforms RS-CMSA, i.e., it achieves a peak ratio of 1.0 faster,  while the reverse is true for the Shubert function, (6 and 8). All these problems are still fully solved by both algorithms.

Problems 13,14, 16, and 18 are all versions of Composition Function 3 (CF3), with increasing problem dimensionality, and contain two niches of Griewank, two of Weierstrass, and two of EF8F2 \cite{CEC2013NichingCompetition}. CF3-2D and 3D (problem 13 and 14) are fully solved for all repetitions. In 5D, RS-CMSA is unable to solve the Weierstrass niches, while {\alg}-AMu occasionally solves them. In 10D, both algorithms achieve the same peak ratio, again being unable to solve the Weierstrass niches. 

Problems 15, 17, 19 and 20 are all versions of CF4 with increasing problem dimensionality. For problem 15, the two Weierstrass niches are almost never solved. In problem 20, the Griewank function can furthermore not be solved by either of the methods.

Interestingly, in one repetition of problem 20, RS-CMSA accidentally discarded a previously detected global optima in the post-processing step. Overall, in the long run, {\alg}-AMu outperforms RS-CMSA.

\label{sec:long_run}

\section{Summary and Discussion}
\label{sec:Discussion}
In this work, we introduced a principled approach of adaptively clustering the search space into promising niches based on {\clusteralg}. Equipped with a restart scheme and various core search algorithms, {\algFull} (\alg) is compared to state-of-the-art EAs designed for MMO. 

{\alg} performs very well on the CEC2013 niching benchmark. Combined with AMaLGaM-Univariate as the core search algorithm, performance is only slightly inferior to the best performing algorithm in the GECCO2017 niching competition, RS-CMSA. {\alg}-AMu achieves an average peak ratio of 0.847 over all 20 problems, compared to 0.856 of RS-CMSA. The choice of core search algorithm does not heavily affect the overall performance of {\alg}.

Using an elitist archive and including it in the population upon restarting further improves the performance of {\alg} on problems with large niches, preventing these niches to be explored multiple times. A drawback is that budget is still spent on re-clustering these niches in the initialization phase. A way to overcome this would likely improve {\alg} further. 

By automatically tuning the restart parameters, we showed that the manually chosen parameters perform well, and performance is robust with respect to many different parameter choices. 

In comparison to NEA2+, {\alg} considers multiple neighbours in the cluster process. The cutting rules are not tuned based on the EA's performance, but each edge is individually considered using the Hill-Valley test. This improves the robustness and quality of clustering, justifying the additionally spent function evaluations. Furthermore, we showed that {\alg} does not significantly improve by replacing the initial sampling algorithm. Standard uniform sampling is therefore preferred, as it is computationally cheaper.

Benchmark problems with a budget-limit do demonstrate the full potential of the tested algorithms. By comparing the performance of {\alg}-AMu with RS-CMSA over time, we showed that {\alg} outperforms RS-CMSA in the long run. However, even for a budget of 1 billion evaluations, neither of the methods are able to detect all global optima for all problems, which suggests limitations in the core search algorithms, in the two-phase approach, or an extremely increased problem complexity for some of the global optima as the dimensionality increases. Re-clustering after a number of generations might help to overcome this. On the other hand, re-clustering might deteriorate performance on problems where it is not necessary, so an adaptive approach may be warranted.

Other improvements to {\alg} might consist of terminating core search algorithms early when it is clear that they are not approaching global optima, or include task-scheduling in order to run each of the core search algorithms based on their probability of finding a new global optimum. 

\section{Conclusion}
\label{sec:conclusion}
In this work, we introduced the {\algFull} ({\alg}), a two-phase niching method, driven by the principled, yet remarkably simple {\clusteralg}. The resulting model-based evolutionary algorithm was found to be surprisingly competitive to state-of-the-art niching methods, and is even superior in the long run on a variety of benchmark problems. We conclude that {\alg} is an interesting new addition to the existing spectrum of evolutionary algorithms for multi-modal optimization, with high potential that may still be further exploited.

\begin{anonsuppress} 
\section{Acknowledgements}
This work is part of the research programme IPPSI-TA with project number 628.006.003, which is financed by the Netherlands Organisation for Scientific Research (NWO) and Elekta. The authors acknowledge the Nijbakker-Morra Stichting for financing a high-performance computing system.
\end{anonsuppress}

\bibliographystyle{ACM-Reference-Format}
\bibliography{Maree_Gecco18} 

\end{document}